\documentclass{IOS-Book-Article}

\usepackage{mathptmx}
\usepackage{soul}\setuldepth{article}
\usepackage{amsmath,amssymb,amsfonts}
\usepackage{algorithm}
\usepackage{graphicx}
\usepackage{textcomp}
\usepackage[dvipsnames,x11names,svgnames]{xcolor}
\usepackage[most]{tcolorbox}

\usepackage{comment}
\usepackage{algpseudocode}
\usepackage[normalem]{ulem}
\usepackage{booktabs}
\usepackage[T1]{fontenc}
\usepackage{todonotes}
\usepackage{tabularx}
\usepackage{multirow}
\usepackage{array}
\usepackage{enumitem}
\usepackage{adjustbox}
\usepackage{ulem}
\usepackage{float}

\usepackage{tcolorbox}
\usepackage{caption}

\usepackage{placeins}

\usepackage{listings}

\usepackage{wrapfig}

\usepackage{booktabs}
\usepackage{array}
\usepackage{caption}

\def\hb{\hbox to 11.5 cm{}}

\begin{document}

\pagestyle{headings}
\def\thepage{}
\begin{frontmatter}              % The preamble begins here.

%\pretitle{Pretitle}
\title{Surprising Effectiveness of Self-Demonstrations in Enhancing Schema-Ontology Mapping with LLMs}

\markboth{}{May 2026\hb}
%\subtitle{Subtitle}

\author[A]{\fnms{Siddhesh} \snm{Thombre}}
% \thanks{Corresponding Author: Siddhesh Thombre, siddhesh.thombre@tcs.com}},
\author[B]{\fnms{Manasi} \snm{Patwardhan}}
and
\author[B]{\fnms{Sunita} \snm{Sarawagi}}

\runningauthor{B.P. Manager et al.}
\address[A]{TCS Research}
\address[B]{IIT Bombay}

\begin{abstract}
Integrating heterogeneous relational databases into a centralized ontology remains a persistent challenge in enterprise knowledge representation, primarily due to semantic heterogeneity, cryptic schema naming, missing metadata, and the abstraction gap between relational schemas and ontological models. Although large language models (LLMs) offer strong semantic reasoning capabilities, we show that directly applying them through one-shot prompting or naïve multi-stage pipelines leads to poor performance for schema–ontology mapping. This paper presents a self-demonstration-driven approach that combines a neuro-symbolic task decomposition with a novel mechanism for automatically generating pattern-guided, dependency-aware demonstrations to address this integration challenge. Our approach incorporates two key strategies to achieve substantial accuracy gains over existing LLM-based schema integration methods: (i) a neuro-symbolic decomposition of the task into cascaded sub-tasks, where symbolic constraints structure the search space and LLMs perform semantic reasoning within each focused sub-task, and (ii) self-generated demonstrations guided by domain-agnostic patterns  to supervise each sub-task. Experiments on three of the most challenging scenarios from the RODI benchmark show that our approach achieves state-of-the-art performance, substantially outperforming ($\sim$25 percentage points F1 improvements) both traditional schema-to-ontology mapping techniques and recent LLM-based schema-to-ontology and schema matching approaches. Ablation studies further reveal the significant benefits of pattern-guided self-demonstrations and the complementary benefits of neuro-symbolic task decomposition.
\end{abstract}

\begin{keyword}
Database-Ontology mapping \sep data integration \sep enterprise search \sep R2RML generation
\end{keyword}
\end{frontmatter}
\markboth{May 2026\hb}{May 2026\hb}

\begin{figure*}[t]
  \centering
   \includegraphics[width=\textwidth]{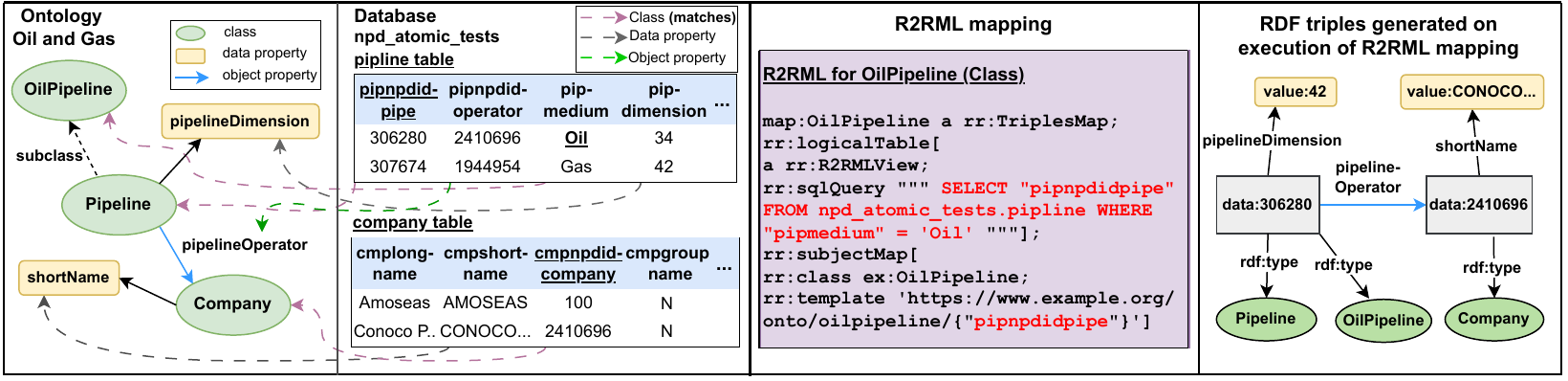}
  \caption{RDB-to-Ontology Mapping Example}
  \label{fig:example mapping}
\end{figure*}

\section{Introduction}

Enterprises increasingly seek to represent organizational knowledge through a single unified ontology so that analytics, search, and reasoning can operate over a semantically consistent enterprise view~\cite{10.14778/3415478.3415512, 10.1007/s11280-025-01355-x,rodriguez2013ontology}. In practice, however, enterprise data is distributed across multiple heterogeneous and independently evolving relational databases. As a result, querying across these sources requires mapping each database schema to a centralized ontology (Example illustrated in Figure \ref{fig:example mapping}).
% Doing this manually is labor-intensive and does not scale.
Doing this manually is a time-consuming, error-prone, and cognitively demanding task at enterprise scale,
% where ontologies 
% often span hundreds of entities, while individual databases can include hundreds of tables and thousands of columns.
given the large number of ontology entities, database tables, and columns involved.

Traditional automated mapping systems~\cite{Pinkel2017RODIBR,Sicilia2016AutoMap4OBDAAG} rely heavily on lexical similarity and hand-crafted templates; consequently, they often miss deeper semantic correspondences, struggle with cryptic or noisy schema elements, and fail when key metadata is incomplete. Large language models (LLMs) are a natural fit for this task because they offer broad semantic reasoning capabilities~\cite{Kojima2022LargeLM, Brown2020LanguageMA,Wei2022EmergentAO, Wei2022ChainOT}
% \manasi{\cite{}} 
that can bridge representational gaps between schemas and ontologies. Yet our early experiments show that directly prompting an LLM to produce complete DB-to-ontology mappings in one shot yields poor results.
When the overall task is organized in a neuro-symbolic manner, where symbolic structure decomposes the task into stages and constrains the search space while LLMs perform semantic reasoning within each stage, the model makes better local decisions and produces better mappings.
However, the benefits yielded by this staged LLM driven approach in zero-shot setting leads to sub-optimal results as inaccuracies at each LLM stage accumulate, leading to a decline in overall performance.
To improve accuracy, we sought to harness the in-context learning (ICL)~\cite{Brown2020} capability of LLMs to adapt to new tasks or domains using demonstrations with a few input-output exemplars. We observe that manually curated demonstrations substantially improved accuracy, but they are labor-intensive and not scalable due to the need for domain and task expertise.%, especially for large-scale mapping tasks. 

Matchmaker~\cite{Seedat2024MatchmakerSL,seedatbootstrapping}, an approach for schema matching, demonstrates the capability of LLMs to automatically generate few-shots for self-improvement. However, extending their approach for our pipeline does not yield us performance on par with the manually curated few-shots. This was mainly because matchmaker's approach tend to generate similar-looking demonstrations, with limited semantic variations in the matches.

This leads to key insight of this work: We allow LLM to generate self-demonstrations while constraining generation using typical reusable, domain-agnostic patterns observed across real data integration cases.
We define pattern families that capture common regularities and guide the LLM to instantiate them,
% as stage-specific
for each stage, to generate self-demonstrations. For example, a class in an ontology matches a DB table following either an exact string-based, syntactic, semantic, domain-based, or categorical-column-value-pair-based match. An object property connecting two classes, in contrast, matches with either the foreign key column of the matched table of the domain class or the columns of the junction table that links the tables matched with the domain and range classes. 
% (detailed patterns are listed in Table \ref{tab:all_patterns}) 
The demonstrations are created in a dependency-aware manner: outputs from earlier stages (such as class/table alignments) are carried forward while generating demonstrations for downstream sub-tasks (e.g., property matching and SQL-view mapping). We show that our staged process of automatic generation and utilization of pattern-grounded, dependency-aware self-demonstrations, which are diverse and less noisy,
%than unconstrained self-generated examples. We show that our self-demonstration strategy
is  effective and is the primary driver of performance gains for LLM driven database to ontology mapping task.
Although our approach invokes LLMs at multiple stages, database to ontology mapping is typically a one‑time integration process whose resulting mappings are reused extensively across downstream queries and applications. As a result, we prioritize accuracy and semantic correctness over execution efficiency or compute cost in enterprise settings.
The key contributions of our approach are summarized as follows:
\begin{itemize}

\item We propose a neuro-symbolic decomposition based approach that combines the strengths of traditional symbolic approaches such as Milan~\cite{Mathur2018MilanAG} with the power of modern LLMs. The database-to-ontology mapping task is systematically decomposed into multiple stages and fine-grained sub-stages, each executed using self-demonstration–guided LLMs.

\item We introduce a pattern-guided context-aware self demonstration method that automatically constructs diverse, stage-specific in-context exemplars, with one-time domain agnostic pattern curation.
% without manual curation.

\item Our approach 
% comprising of self-demonstrations and task decomposition
achieves state-of-the-art results on the three most challenging scenarios from the RODI benchmark ~\cite{Pinkel2017RODIBR} (geographical, conference and oil and gas), demonstrating substantial improvements over direct prompting and prior automated baselines ($\sim$25 percentage points in F1 score)
and even slightly better results than manually curated few-shots.
\item We show via detailed ablations that our approach benefit from neuro-symbolic task decomposition and self demonstrations.
\end{itemize}

\begin{figure*}[t]
  \centering
\includegraphics[
  width=0.99\textwidth,
  height=0.096\textheight
%   keepaspectratio
]{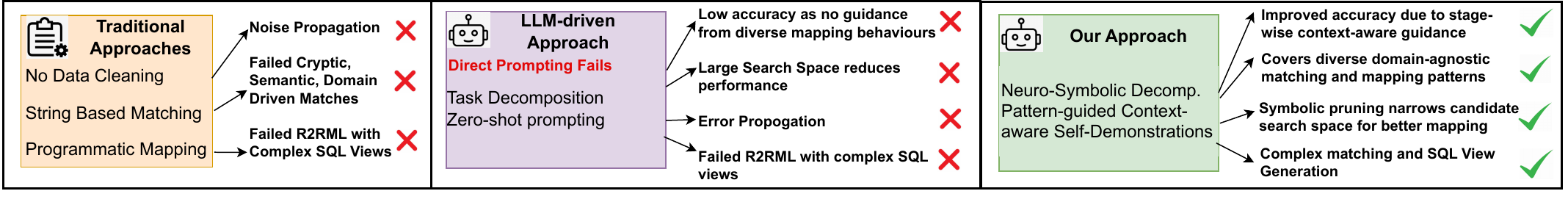}
  \caption{Motivation behind Our Approach of DB-to-Ontology Mapping}
  \label{fig:mapping}
\end{figure*}

\section{Related Work}
% \paragraph{DB-Ontology Mapping.}
\textbf{DB-Ontology Mapping.}
Early systems such as BootOX~\cite{JimnezRuiz2015BootOXPM}, Ontop~\cite{Calvanese2016OntopAS}, MIRROR~\cite{Medeiros2015MIRRORAR}, and D2RQ~\cite{Oldakowski2011D2RQP} derive a bootstrapped ontology from a database and then try to map this ontology with the schema, which is not a realistic setting. Subsequent tools like COMA++~\cite{Aumller2005SchemaAO}, IncMap~\cite{Pinkel2017IncMapAJ}, A4MO~\cite{Sicilia2016AutoMap4OBDAAG}, and MILAN~\cite{Mathur2018MilanAG} rely mainly on lexical/structural similarity with heuristic pruning and template-based mapping. These approaches degrade under cryptic nomenclature, incomplete metadata (e.g., missing FKs), and when semantic correspondences transcend string similarity. Recent LLM-based methods (Table-to-KG~\cite{Vandemoortele2024ScalableTG} and LLM4VKG~\cite{Xiao2025LLM4VKGLL}) introduce learned semantic matching with LLMs but often stop at alignment, keep mapping template-driven, or do not provide targeted, stage-specific guidance to the LLM. 
% Our study shows that pattern-guided self-demonstrations supply exactly that missing guidance, a simple but impactful ingredient to make LLMs effective on this task.\\
In contrast, our approach combines symbolic decomposition and pruning with LLM-based semantic reasoning, and further provides targeted, stage-specific guidance through pattern-guided self-demonstrations.\\
\textbf{Schema Matching.}
In the DB-to-ontology mapping task, the initial correspondence identification step closely resembles schema matching.
LLM-based schema matching commonly follows “retrieve candidates then re-rank with an LLM,” as in ReMatch~\cite{Sheetrit2024ReMatchRE} and Magneto~\cite{Liu2024MagnetoCS}. Matchmaker~\cite{Seedat2024MatchmakerSL,seedatbootstrapping} further explores automated few-shot self-improvement. Yet schema matching is typically simpler than DB–ontology mapping given the abstraction and granularity mismatch between schemas and ontologies. Matchmaker's few-shot generation approach when adapted to our setting leads to sub-optimal results. In our setting, automated few-shots help only when they are diverse, scenario-specific, and 
% stage-aware,
dependency-aware,
characteristics we enforce via self-demonstrations.\\
\textit{Takeaway}.
% Prior literature indicates that LLMs can capture semantics beyond strings, but it underplays \emph{how to feed LLMs the right in-context signals}. We operationalize this through \emph{pattern-guided self-demonstrations}, automatically synthesized, diverse, and scenario-specific exemplars that consistently steer the model across mapping stages.
Prior literature indicates that LLMs can capture semantics beyond strings, but it underplays how to combine semantic reasoning with symbolic decomposition and how to feed LLMs the right in-context signals. We operationalize this through a neuro-symbolic design with pattern-guided self-demonstrations, automatically synthesized, diverse, and scenario-specific exemplars that steer the model across mapping stages.

% \section{Our Approach: \sysname}
\section{Background and  Problem Definition}

We formalize the problem and recall essential background on ontology structure and schema‑to‑ontology mappings.
We are given as input: (i) an Ontology $\mathcal{O}$ consisting of a set-of classes $C$ such as \texttt{Pipeline} in Figure \ref{fig:example mapping}, subclasses  $C_S \rightarrow C$ (e.g. \texttt{OilPipeline}) for a subset of classes $C$, data properties $D_C$ (e.g. \texttt{pipelineDimension}) with domain as a class (e.g. \texttt{Pipeline}) and range as its data type (such as `float'), and object properties $O_{C_D,C_R}$ (e.g. \texttt{pipelineOperator}) linking two classes, a domain class $C_D$ (e.g. \texttt{Pipeline}) and a range class $C_R$ (e.g. \texttt{Company}), and (ii) Relational Database schema $\mathcal{D}$ consisting of entities such as tables $T$ (e.g. \texttt{pipeline}), columns belonging to those tables $C_T$ (e.g. \texttt{pipmedium}), and data values of those columns $V_{C_T}$ (e.g. `\texttt{Oil}').

An Ontology-DB mapping entails associating ontology concepts (class, data properties and object properties) with corresponding database entities (tables, columns).
% For example, an ontology class $c$ may map to primary key columns of zero, one, or more DB tables $T_c$ or a table subset with column $c_{T_c}$ filtered with a value $v_{c_{T_c}}$.
Typical regularities include: classes map to table primary keys or filtered table subsets via categorical columns; subclasses map to column–value filters in or adjacent to superclass tables; data properties map to columns in class tables or one-hop neighbors; object properties maps to foreign keys or junction-table keys connecting domain and range.
% , execution of which leads to the migration of data from relation databases to centralized ontology. 

A snippet of such a mapping task between a subset of oil and gas ontology made for the Norwegian Petroleum Directorate (NPD) and a relational DB is shown in Figure \ref{fig:example mapping}. Mappings are expressed in R2RML (Relational to Resource Description Framework Mapping Language)~\cite{DasSundaraCyganiak2012}.
Our goal is to generate a set of R2RML mappings consisting of SQL views, with subject and object columns as illustrated in Figure \ref{fig:example mapping} such that execution of the R2RML mapping leads to RDF triples, populating $\mathcal{O}$ with the data in $\mathcal{D}$.

\begin{figure*}[!t]
  \centering
  \includegraphics[width=\textwidth]{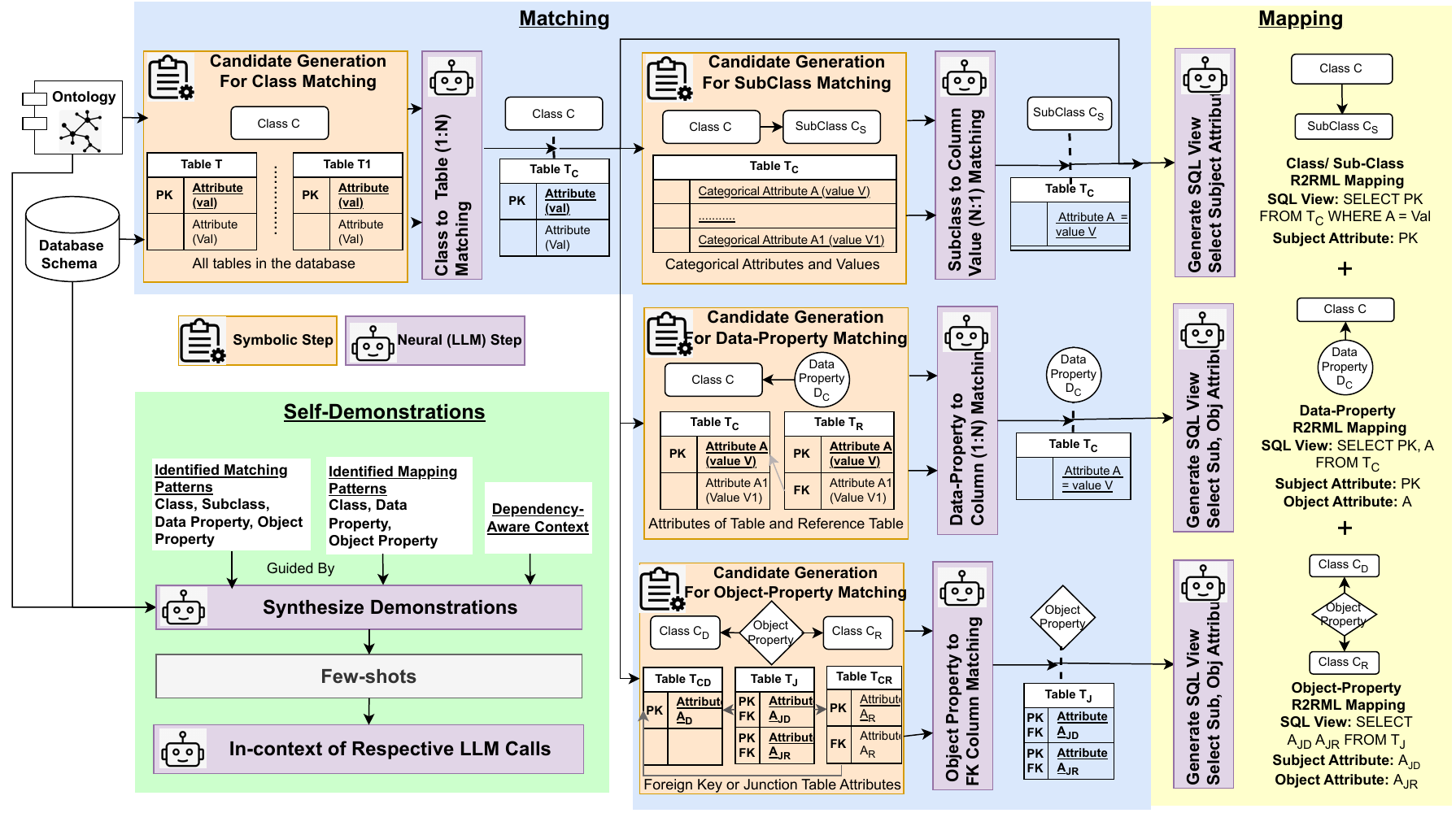}
  \caption{Overview of Self-Demonstration Driven Mapping Approach. Decomposition: (i) \textcolor{LightBlue1}{Matching}, (ii) \textcolor{Yellow}{Mapping} and \textcolor{LightGreen}{Self-Demonstrations}. $T_C$: Table matched with class $C$, $C_S$: Subclass of Class $C$, $D_C$: Data-property of Class $C$, $T_K$: Table referring to $T_C$, $T_J$: Junction Table }
  \label{fig:algo}
\end{figure*}

\section{Self-Demonstration Driven Approach}

\subsection{Overview of Neuro-Symbolic Decomposition}

Motivated by the observation that despite long context support, current LLMs struggle to produce accurate R2RML mappings in a single, one‑shot pass, we decompose the DB-to-Ontology mapping task into two stages: 
(i) Matching: Identifying the most relevant target entities in the DB schema $\mathcal{D}$ that correspond to a source concept in the ontology $\mathcal{O}$,
(ii) Mapping: Constructing R2RML mappings for each ontology concept $\mathcal{O}$ by generating SQL views over matched schema entities. 
We infer missing foreign keys in $\mathcal{D}$ and  domain or range declarations in $\mathcal{O}$ when such metadata is absent, before serving them as an input to the above stages. 
% Both stages are further divided into sub-stages for alignment of each of the ontology concepts, preserving their dependencies. 
Both stages are further divided into sub-stages for aligning ontology concepts, preserving their dependencies. 
Similar to Milan~\cite{Mathur2018MilanAG}, we address
% the challenge of
the vast schema–ontology search space by performing matching in a top‑down, stage-wise manner, progressively pruning candidates using symbolic constraints. We decompose the overall task into sub‑stages based on ontology entity types and construct focused candidate sets using rule-based reasoning, substantially reducing the number of database entities taken into consideration as candidates compared to full database schema. 
Unlike Milan, we tightly integrate this symbolic decomposition with neural component by invoking LLMs at each sub‑stage, equipped with carefully scoped contextual information. Operating over these focused candidate sets allows the LLM to reason within a constrained and relevant context, leading to more accurate and robust mappings ~\cite{Du2025ContextLA, Hsieh2024RULERWT, Shi2023LargeLM, Yang2025HowIL}. 

(i)\textit{Matching.} Matching proceeds in four sub-stages in the order: class, subclass, data property, and object property.
% For example, in the NPD database--ontology pair from the RODI benchmark (see Section~\ref{sec:dataset}), symbolic pruning reduces the candidate search space by approximately 95\% for data property matching and 85\% for object property matching stages, thereby enabling more focused and accurate matching.
Class matching is a two-step process that first selects the best matching table, then additional matching tables if present, are identified. Subclass matching is invoked only when the superclass is matched and searches categorical column–value pairs in the superclass tables and in one hop related tables. Data property matching draws candidates from tables matched to the domain class and from one hop related tables. Object property matching runs only when both domain and range classes are matched, considering candidates from the domain and range tables, one hop related tables, and keys of the junction table, when applicable.
For example, in the NPD database--ontology pair from the RODI benchmark (see Section~\ref{sec:dataset}), symbolic pruning reduces the candidate search space by approximately 95\% for data property matching and 85\% for object property matching stages, thereby enabling more focused and accurate matching.

(ii)\textit{Mapping.} Mapping is decomposed into class, data property, and object property mapping. For each ontology concept, the model generates a SQL view and identifies the subject and object columns required by R2RML using %the DB schema context, stage specific self-demonstrations, and 
the  matches obtained in the matching stage for that concept. Class mapping selects subject identifiers from primary keys of the matched class tables. For subclasses, the mapping involves generating SQL filter conditions (e.g., category predicates) that distinguish subclasses when they match to a shared table.
Data property mapping combines the matched value column with the identifiers of the domain class, while object property mapping links identifiers of the domain and range classes through the matched relation columns, including junction table paths when required. Per concept outputs are aggregated into the final R2RML mapping file. 
The matching and mapping prompts, along with the symbolic steps, are provided in the repository\footnote{https://anonymous.4open.science/r/Schema-Ontology-Mapper-4BC1/}. Figure~\ref{fig:algo} provides an overview of this decomposition.

This sequential stage-wise structure improves mapping quality but the decisive gains arise when each stage is augmented with self‑generated, pattern‑guided demonstrations. These demonstrations are instantiated on the input DB–ontology pair and supplied in-context at the point of use,
% steering the model toward higher fidelity decisions.
improving decision fidelity across the above stages.

\subsection{Why Pattern-guided Context-aware Self-demonstrations.}
% Providing Manually curated stage-wise few-shots substantially improve the mapping quality but curation is challenging and does not scale. Automated selection methods from schema matching (Matchmaker) tend to bias toward “easy” exemplars and miss the breadth of mapping phenomena. We instead \emph{generate} demonstrations with the LLM and \emph{constrain} them with a compact, reusable library of domain-agnostic patterns observed across DB–ontology mapping.
Our initial zero-shot LLM experiments for the matching and mapping sub-tasks yield sub-optimal results. We observe that manually curated exemplars for In-Context Learning (ICL)  lead  to substantial performance improvements. However, this manual curation is labor-intensive and not scalable for distinct DB-Ontology pairs since it requires domain and task expertise. Hence, we seek methods of auto-generation of few-shots.% by harnessing the LLM in creative ways.

One of the Schema Matching approaches, Matchmaker~\cite{Seedat2024MatchmakerSL,seedatbootstrapping}, utilized automatic few-shot generation for self-improvement in schema-to-schema matching. We observe that adapting Matchmaker for schema-to-ontology mapping,  often produce exemplars concentrated around a narrow style of semantic similarity, with limited coverage of the broader matching and mapping behaviors required for DB-to-ontology integration (Details in Section \ref{sec:research questions}).
Database to ontology mapping problem is  fundamentally harder than schema matching because databases and ontologies represent knowledge at different levels of
abstraction and granularity. Relational schemas are designed for efficient storage and query execution, and therefore expose low-level, fine-grained structures such as tables and columns. Ontologies, in contrast, encode high-level domain concepts, constraints, and relationships. This mismatch is amplified in enterprise settings where naming conventions vary across systems, semantically related entities use dissimilar labels, and similarly named entities may carry different meanings depending on context. This leads to distinct scenarios or patterns of this mapping task, with much higher variations than schema matching. Thus, diversity in few-shots is critical to cover varied matching and mapping scenarios~\cite{Kapuriya2025ExploringTR,Thorpe2024DuboSQLDR}.  
Further, each sub-stage of schema‑to‑ontology mapping task relies on information produced by previous ones and thus, synthesizing demonstrations for any sub-stage must incorporate dependency‑aware context (i.e, the outputs from earlier stages) to ensure coherence and correctness of the demonstrations. Thus, unlike schema‑matching pipelines, the schema‑to‑ontology mapping stages %in our approach 
exhibit strong sequential dependencies to avoid error-propagation through  stages. %Because This need for stage‑conditioned exemplars further complicates demonstration generation and makes naive adaptation of existing automatic few‑shot methods insufficient.

To avoid manual intervention but at the same time retain the quality and diversity of few-shots, we devise the self-demonstrations method to auto-generate few-shots for any given DB-ontology pair with minimal one-time guidance using a reusable library of domain-agnostic patterns and dependency-aware context.
% \textcolor{blue}{Furthermore, the sequential and inter‑dependent nature of our approach requires incorporating dependency‑aware context (i.e., outputs from preceding stages), when synthesizing demonstrations for any given stage. }
% Additionally, the sequential dependencies across stages necessitate incorporating context from earlier stage outputs when generating demonstrations for any subsequent stage.

\begin{wrapfigure}{r}{0.55\textwidth}
  \centering
  \vspace{-0.5\baselineskip} % optional: pull it up a bit
  \includegraphics[width=0.55\textwidth]{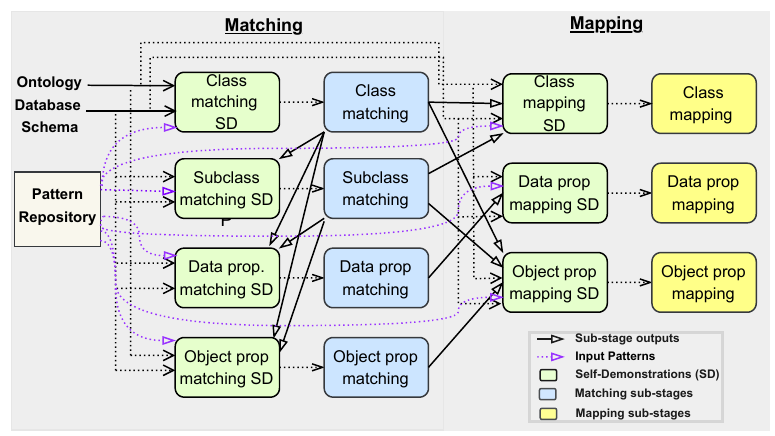}
  \caption{Dependency-Aware context for Self-Demonstrations}
  \label{fig:dependency_aware_SD}
  \vspace{-0.5\baselineskip} % optional: reduce space below
\end{wrapfigure}

\subsection{Synthesis of Self-Demonstrations}\label{sec:self-demos}
% Table~\ref{tab:all_patterns} captures the patterns used to instantiate demonstrations. 
We provide the LLM with an exhaustive set of predefined domain-agnostic input patterns that demonstrate distinct strategies of matching and mapping (patterns are provided in the repository). With these patterns in context, we task the LLM to instantiate them on the given DB-ontology pair. Thus, starting from generic patterns, we generate a comprehensive and diverse set of few-shots for any given DB-ontology pair.  With this approach, we ensure that the generated few-shots cover a wide array of matching and mapping cases. 

A pattern is a template that captures a matching/mapping behaviour. The pattern library encodes recurring correspondence behaviors seen across DB-ontology integration tasks: lexical and semantic alignment patterns (e.g., exact, paraphrased, or abbreviation-based matches), structural patterns (e.g., object properties realized through foreign keys or junction tables), and value-constrained patterns (e.g., subclasses represented as filtered subsets based on categorical column values). We also include mapping-side patterns for SQL view construction, such as projection-only mappings, join-based mappings, and mappings requiring value normalization or conditional filtering. This pattern-oriented organization makes each demonstration explanatory rather than merely illustrative: the few-shot not only states the final match or SQL fragment, but also reflects the underlying pattern that justifies it. Patterns also act as a controllable diversity mechanism. During demonstration synthesis, we explicitly prompt the LLM to cover different pattern families instead of selecting random examples. As a result, the final in-context set spans both straightforward and complex behaviors, improving robustness when the target concept exhibits ambiguous naming, implicit relations, or non-trivial SQL transformations.
% \textcolor{blue}{For instance, one of the class‑matching patterns involves class matching a categorical column. This pattern represents the common scenario where a class corresponds to a subset of a table’s rows, identifiable by filtering the table on a categorical column with a specific value.}

Few-shots are generated for each matching and mapping sub-stage  sequentially by providing stage specific patterns (prompt is provided in repository).
For each sub-stage, LLM leverages the outputs of the relevant prior stage(s) to further improve the  quality of the demonstrations. For example, for generation of few-shots for subclass, data and object property matching step, class matching output serves as additional input. Whereas for data and object property mapping, the output of data and object property matching serve as an input (details are provided in Figure \ref{fig:dependency_aware_SD}). 

This dependency-aware context improves demonstration quality for two reasons. First, it guides the LLM to focus on  candidates that are structurally consistent with earlier decisions (e.g., data-property examples are grounded in tables already selected for the domain class), which lowers semantically plausible but invalid matches.  
For example, consider the data property \texttt{hydrocarbonType}, whose domain is the class \texttt{Discovery} and whose range is string. From the earlier class matching stage, the class Discovery has been aligned with the \texttt{discovery} table. During the self demonstration generation step for property matching, this prior alignment is included as contextual information. By making the class to table correspondence explicit, the context directs the LLM’s attention toward columns associated with the \texttt{discovery} table, leading it to focus on candidates such as \texttt{discovery.dshctype} rather than exploring other semantically plausible but structurally less aligned columns such as \texttt{dscarea.dshctype} and \texttt{fldarea.dshctype}.

Second, because established upstream decisions (e.g., class-table alignments) are passed into downstream demonstration construction, demonstrations across downstream tasks (e.g., data property alignment / mapping) become more consistent and correct.  
For example, consider the data property \lstinline|wellboreStratumBottomDepth|, whose domain is the class \lstinline|WellboreStratum| and whose range is decimal. In an earlier class‑matching stage, the \lstinline|WellboreStratum| class is aligned with the \lstinline|wellbore_formation_top|, \texttt{strat\_litho\allowbreak\_wellbore} tables using richer contextual signals, giving these class table alignments a higher likelihood of being correct. When these higher confidence upstream decisions are incorporated as contextual input during the self demonstration generation step for property matching, they provide reliable structural anchors for downstream reasoning. As a result, the LLM constructs property‑level demonstrations under the assumption that \texttt{wellboreStratumBottomDepth} should be interpreted within the context of the \texttt{wellbore\_formation\_top} and \lstinline|strat_litho_wellbore| tables, leading to consistent demonstrations that align the property with \texttt{wellbore\_formation\_top.lsubottomde-\\pth} and \texttt{strat\_litho\_wellbore.lsubottomdepth}. In contrast, without incorporating these upstream alignments, the LLM may either arbitrarily select a single table or treat multiple semantically plausible columns as equally valid, increasing the likelihood of generating inconsistent or incorrect property‑level demonstrations.
% Thus, the generation of few-shots in this sequential manner with dependency-aware context leads to better quality and consistent examples, avoiding error-propagation through the sub-stages. 
% Thus, dependency-aware few-shot generation yields more consistent examples and limits error propagation across sub-stages.
Thus, dependency-aware few-shot generation improves example quality and consistency while limiting error propagation across sub-stages.
% Figure \ref{fig:dependency_aware_SD} provides the details about the context utilized while synthesizing self-demonstrations within different stages of our approach.
Figure \ref{fig:dependency_aware_SD} summarizes the context used for self-demonstration synthesis across stages.

In manual few‑shot curation, an expert selects demonstrations, each corresponding to a distinct matching or mapping pattern in the DB-ontology pair. Identifying representative examples for new pairs is challenging and time‑consuming. In contrast, our automatic approach instantiates at most one example per applicable pattern. 
% The number of few‑shot examples is identical in both settings (one per pattern).
Both settings use the same number of few-shot examples (one per pattern).
% We use the same fixed domain-agnostic pattern set across all input DB--ontology pairs; new domains require no mandatory expert-authored templates, except optional additions for rare missing patterns.
We use the same fixed domain-agnostic pattern set across all input DB--ontology pairs; new domains require no expert-authored templates, though experts may optionally add rare missing patterns.
% The full set of patterns is documented in the project’s GitHub repository.

\section{Experimentation}
% \subsection{Dataset}
\subsection{RODI Benchmark}
\label{sec:dataset}

\begin{wraptable}{r}{0.55\columnwidth}
  \vspace{-0.8\baselineskip}
  \centering
  \scriptsize
  \setlength{\tabcolsep}{2.5pt}
  \renewcommand{\arraystretch}{1.15}

  \begin{tabular}{@{}lrrrrrr@{}}
    \toprule
    \textbf{Name} &
    \textbf{Classes} &
    \textbf{Props.} &
    \shortstack{\textbf{}\\\textbf{Tables}} &
    \shortstack{\textbf{}\\\textbf{Cols.}} &
    \textbf{FKs} &
    \textbf{Queries} \\
    \midrule
    Conference & 23  & 77  & 66 & 125  & --  & 39  \\
    Mondial    & 49  & 71  & 42 & 160  & 60  & 50  \\
    Npd        & 300 & 350 & 70 & 1000 & 100 & 439 \\
    \bottomrule
  \end{tabular}

  \caption{RODI Scenarios}
  \label{tab:stat}
  \vspace{-0.8\baselineskip}
\end{wraptable}

We use RODI (Relational-to-Ontology Data Integration)~\cite{Pinkel2017RODIBR}, benchmark to test the R2RML generation task. This benchmark has 18 test scenarios belonging to three domains. %Out of 18 scenario 15 belongs to conference domain, where the database schema is synthesized based on distinct  conference ontology (CMT, SigKDD, etc) and then various scenarios are created by adjusting the mappings, restructuring, de-normalizing, removing the foreign key information from the synthesized databases or considering cross-domain mapping, where the task is to map a database synthesized for a conference to the ontology of another conference. 
Each scenario consists of a DB, an ontology and set of evaluation queries to test expected results. %The evaluation queries comprise of pairs of test SPARQL query and reference SQL query. The test query runs against the RDF data that results after execution of generated R2RML file with an approach under consideration. The reference SQL query is directly evaluated by RODI against the SQL database. The results are compared for each query pair to compute per-query F-measure to determine the quality of generated mappings for each scenario.  

We experiment with one representative scenario from each domain, viz., Mondial for Geographical Data, NPD-Atomic for oil and gas, and Conference-no-FK, the most challenging scenario within each domain~\cite{Mathur2018MilanAG}. Table \ref{tab:stat} presents the statistics for the DB entities and ontology concepts of these scenarios.  These span large DB schema and ontology of significant complexity, providing a realistic simulation of enterprise-level integration challenges. The Conference-no-FK scenario is especially difficult because DB schema lacks foreign key definitions, thus introducing ambiguity and noise into the task. For example, \texttt{was\_a\_program\_committee\_of} column of the \texttt{Committee} table should ideally be a foreign key of the  \texttt{Conference\_volume} table, but this relationship is missing in RODI.
The mapping of data and object properties of the Mondial scenario exhibits a high degree of complexity, involving domain and range definitions, with logical combinations of multiple classes. For example, the domain class of \texttt{type} data property is defined as a complex expression involving unions, intersections and negations of classes, such as "((\texttt{GeographicalThing} or \texttt{Membership}) and (not (\texttt{City})) and (not (\texttt{Continent})) and (not (\texttt{Estuary})) and (not (\texttt{River})) and (not (\texttt{Source}))". It also involves object properties sharing identical domain and range classes. For example, the object properties \texttt{neighbor} and \texttt{dependentOf} have the same domain and range class as \texttt{Country}. 
The NPD-Atomic scenario represents the most complex case with very large ontology and DB schema. This scenario is characterized by a large number of one-to-many (1:n) mapping and individual ontology concepts must often be mapped to a union of several DB entities, to retrieve complete and correct results.  For example, the mapping of \texttt{productionYear} data property is defined as a union over columns of 5 different tables.\\
\textbf{Metric}:
We adopt the evaluation framework of the RODI benchmark. For each scenario a series of query pairs test a range of mapping challenges which are the result of distinct naming conventions of DB and ontology, structural conflicts caused by process of DB normalization, varied implementations of class hierarchies in an ontology, or semantic heterogeneity stemming from the conflict between a database's closed-world and an ontology's open-world assumption (object-relational gap). Every query pair consists of (i) a SPARQL query (test query) that runs against the RDF ontology data, resulting after execution of the generated R2RML, and (ii) a semantically equivalent SQL query (reference query) evaluated against the corresponding DB. For each query pair, the local per-query F-measure is computed by comparing reference tuples (SQL execution results) with result tuples (SPARQL execution results).
The overall score reported for each scenario is the average of F-measures across test queries. 
More details are provided in~\cite{Pinkel2017RODIBR}.
We additionally report an average score (Avg), defined as the mean of F‑measures computed over all queries across the selected test scenarios.

\subsection{Baselines}\label{sec:base}
Following approaches serve as our baselines: (i) traditional top-performing DB-to-Ontology mapping (Milan~\cite{Mathur2018MilanAG} and A4MO~\cite{Sicilia2016AutoMap4OBDAAG}) (ii) LLM-based DB-to-Ontology mapping (LLM4VKG~\cite{Xiao2025LLM4VKGLL} and table-to-KG~\cite{Vandemoortele2024ScalableTG}) and (iii) LLM-based state-of-the-art schema matching (Magneto~\cite{Liu2024MagnetoCS}).
% \subsubsection{AM4O}
\textbf{AM4O}:
Automap~\cite{Sicilia2016AutoMap4OBDAAG} converts the DB schema into a putative ontology, where tables become classes, columns become data properties, and foreign keys become object properties. 
The classes and data properties of the putative and target ontology are matched using string similarity metrics. The alignment extension step refines these initial matches by analyzing the structural relationships between the DB and ontology concepts. 
 Finally, % R2RML mappings are generated programmatically based on the identified correspondences. %, with each correspondence yielding a triple map in R2RML syntax.
SQL queries for logical tables are programmatically derived using the DB paths identified during the alignment extension step. 
Automap's alignment extension requires defined foreign keys, rendering it ineffective when absent. String similarity struggles with cryptic names or complex domains, and initial matching errors can propagate to subsequent alignment extensions.
We use the results provided in~\cite{Pinkel2017RODIBR}.
\textbf{MILAN}:
MILAN~\cite{Mathur2018MilanAG}  uses %multistage algorithm to detect class-table, data property-column and object property-referential integrity correspondences. These  correspondences
uses Levenstein distance \footnote{https://en.wikipedia.org/wiki/Levenshtein\_distance} %\textcolor{red}{\cite{}}
to match DB-ontology correspondences
and to identify optimal matches using combinatorial optimization. 
MILAN also generates the R2RML mappings programmatically using a set of predefined templates for the SQL views.
It is currently the state-of-the-art traditional approach for the DB-to-ontology mapping task. We use the results provided in~\cite{Pinkel2017RODIBR}.
\textbf{Table-To-KG}:
Table-to-KG~\cite{Vandemoortele2024ScalableTG}  uses LLMs to match table columns with semantic concepts in a knowledge graph. For each DB column, the top-k (k=50) ontology concepts are ranked with the highest cosine similarity of text embeddings. 
We generate the embeddings with Gemini-embedding-001\footnote{https://developers.googleblog.com/en/gemini-embedding-available-gemini-api/}. The top-k matches are re-ranked with an LLM and a sliding window approach (step size = 10, window size = 20). %The final completion stage then employs an LLM to rank the top-5 matching ontology entities from 30 re-ranked candidates. 
Chain-of-Thought with self-consistency~\cite{Wang2022SelfConsistencyIC} (n = 3 samples) is used to retrieve multiple rankings, which are subsequently fused by reciprocal rank fusion to select the top-ranked match. Although the original work uses Llama-3-70B\footnote{https://huggingface.co/meta-llama/Meta-Llama-3-70B} for re-ranking and GPT-4o
% \footnote{https://platform.openai.com/docs/models/gpt-4o}
for completion, for a fair comparison, we use Gemini-2.0-flash for all stages. %Though the original approach uses descriptions of ontology concepts as inputs, we skip the same due to unavailability of such descriptions in the RODI benchmark. 
Also, this method only provides matching, to obtain mapping, we use the programmatic mapping approach like Milan.
\textbf{LLM4VKG}:
LLM4VKG~\cite{Xiao2025LLM4VKGLL} is a recent LLM-based approach to construct Virtual Knowledge Graphs (VKGs), consisting of DB, ontology and mappings. %In addition to generating mappings, LLM4VKG extends the ontology to include database concepts that are not present in it. 
%The approach involves two steps: (i) Mapping Pattern Recognition, which converts the database schema into a graph and identifies mapping pattern instances using SPARQL queries that guide mapping generation; and (ii) Ontology Completion and Mapping Bootstrapping, which leverages LLMs and identified mapping pattern instances to complete the ontology and generate mappings.
It performs the alignment with three modules. Retriever finds semantically similar ontology candidates using a sentence similarity model. Matcher refines matches and assigns a matching degree (High, Medium, Low) via an LLM. Finally, Namer generates new ontology terms when no suitable match exists. 
As a baseline, we use the results provided in~\cite{Xiao2025LLM4VKGLL}.
\textbf{Magneto}:
Magneto~\cite{Liu2024MagnetoCS} is the state-of-the-art approach that leverages LLMs for the schema matching task, with the assumption of having a single table in the source and the target schema. To use this approach as one of the baselines, we customize it for matching ontology concepts to DB entities and execute mapping with Milan's programmatic approach.  For each concept in the ontology, we rank all entities in the DB using a score provided by a retriever (MPNet~\cite{Song2020MPNetMA}), performing a semantic match. Then, for each concept in the ontology,  the top-k ranked DB entities are re-ranked by prompting the LLM. We customize their prompt%(illustrated in Figure \ref{fig:Magneto LLM-reranking prompt})
, originally designed for schema matching to schema-ontology matching. They use GPT-4o-mini \footnote{https://platform.openai.com/docs/models/gpt-4o-mini} for LLM re-ranking. However, for a fair comparison, we keep the LLMs consistent with those we use for our approach.
\textbf{Matchmaker}:
Matchmaker~\cite{Seedat2024MatchmakerSL,seedatbootstrapping} is an LLM-based approach for schema matching. It involves a multi-step pipeline comprising candidate generation (using semantic retrieval and LLM reasoning), LLM-based candidate refinement, and confidence scoring. 
A key feature of Matchmaker is LLM self-improvement through automated in-context example selection. %, with no access to the ground-truth mappings. 
It identifies a set of `easy'  (having a semantic similarity score > 0.95 with target schema entities) and `challenging' (exhibiting the lowest semantic matches) source schema entities. The complete matching pipeline is then executed for each identified entity, with all intermediate outputs stored. An LLM evaluator assesses the final target schema entity match, assigning a relevance score from 0 to 5. The intermediate traces from the top-n source entities with the highest evaluation scores are selected as in-context examples. % This automatic selection of in-context examples is analogous to the self-demonstration stage within our approach. 
To assess the effectiveness of our self-demonstration generation strategy, we compare it with Matchmaker's few-shot example selection strategy adopted and integrated in our pipeline. Specifically, for each stage in our approach, we obtain DB schema matches under zero-shot conditions, for ontology concepts selected using Matchmaker's method. These results are then evaluated by an LLM evaluator, and the top-n scoring ontology concepts and their matches are chosen as in-context examples for each stage.

\subsection{Results and Discussion}
\label{sec:research questions}

\begin{wraptable}{r}{0.58\textwidth}
    \vspace{-1\baselineskip}
    \centering
    \small
    \renewcommand{\arraystretch}{1}

    % Make the first column wrap; keep numeric columns compact and centered
    \begin{tabularx}{0.58\textwidth}{>{\raggedright\arraybackslash}X c c c c}
    \\
        \toprule
        \textbf{Scenario} & \textbf{Conf} & \textbf{Mondial} & \textbf{NPD} & \textbf{Avg} \\
        \midrule

        \multicolumn{5}{l}{\textbf{Traditional Baselines}} \\
        \midrule
        AM4O~\cite{Sicilia2016AutoMap4OBDAAG} & 0.41 & 0.44 & 0.23 & 0.26 \\
        Milan~\cite{Mathur2018MilanAG}        & 0.46 &  -   & 0.30 & 0.31 \\
        \midrule

        \multicolumn{5}{l}{\textbf{LLM Driven Baselines}} \\
        \midrule
        LLM4VKG~\cite{Xiao2025LLM4VKGLL}                     & 0.51 & 0.18 & 0.19 & 0.21 \\
        Table-to-KG~\cite{Vandemoortele2024ScalableTG}\#      & 0.33 & 0.10 & 0.17 & 0.18 \\
        Magneto~\cite{Liu2024MagnetoCS}\#                     & 0.41 & 0.28 & 0.21 & 0.23 \\
        \midrule

        \multicolumn{5}{l}{\textbf{LLM Driven Demonstration Based Baselines}} \\
        \midrule
        Matchmaker~\cite{Seedat2024MatchmakerSL, seedatbootstrapping} & 0.56 & 0.38 & 0.19 & 0.24 \\
        Manually Curated FS                                           & 0.72 & 0.82 & 0.48 & 0.53 \\
        \midrule
        \textbf{Our Approach} & \textbf{0.79} & \textbf{0.79} & \textbf{0.51} & \textbf{0.56} \\
        \midrule

        \multicolumn{5}{l}{\textbf{Ablations}} \\
        \midrule
        - \textit{Self-Demonstrations (FS)}           & 0.54 & 0.46 & 0.36 & 0.38 \\
        - \textit{Matching}                           & 0.51 & 0.50 & 0.36 & 0.38 \\
        - \textit{Mapping}\#                          & 0.62 & 0.36 & 0.43 & 0.44 \\
        - \textit{Matching+Mapping}$\dagger$          & 0.46 & 0.08 & 0.07 & 0.10 \\
        \bottomrule
    \end{tabularx}

    \caption{Results on RODI Benchmark with Gemini-2.0-Flash. Conf: Conference-no-fk; NPD: Npd-Atomic; FS: Few-Shot; \#: With programmatic mapping; $\dagger$: No Decomposition.}
    \label{tab:result}
    \vspace{-1\baselineskip}
\end{wraptable}

\textbf{RQ1: How do LLM-driven approaches compare with traditional approaches?}
Direct single-pass LLM prompting is insufficient for obtaining complete DB-to-ontology mapping, as evidenced by the \textit{-Matching+Mapping} ablation in Table~\ref{tab:result}, which achieves only 0.10 average F1 score and underscores the combined challenge of semantic alignment and R2RML mapping generation when attempted jointly in a one‑shot setting.
% highlights the joint difficulty of semantic alignment and R2RML mapping generation in an end-to-end one-shot setting. 
In comparison, traditional baselines such as Milan and A4MO demonstrate better performance (Table \ref{tab:result}) but remain constrained by dependence on lexical similarity and predefined mapping templates.

Prior LLM-driven systems do not outperform the strongest traditional baseline. Table~\ref{tab:result} shows that LLM4VKG (0.21), Table-to-KG (0.18), and Magneto (0.23) remain below Milan (0.31). This pattern indicates that simply introducing an LLM is not enough, performance depends on how the LLM is embedded in the pipeline.

The observed gap with Milan stems from limitations in earlier LLM‑based approaches. Without decomposing the mapping task into sequential symbolic sub‑stages that progressively constrain candidate sets using rule‑based reasoning, the model faces a large, weakly constrained search space, making it more prone to imprecise or incorrect mappings.
In addition, the lack of stage specific supervision in the form of high quality few shot exemplars provides insufficient guidance on the expected behavior of different mapping sub-types. 
% Element wise matching without sufficient contextualization degrades performance, for example, when the model attempts to align a property of a given class without knowledge of the other properties of that class, the table matched to the class, or candidate columns along with representative example values \manasi{incomplete sentence}. 
Element wise matching without sufficient contextualization further degrades performance; for example, when the model attempts to align a property of a given class without knowledge of other properties of that class, the table matched to the class, or relevant candidate columns along with representative example values, it may select a plausible but incorrect candidate due to the absence of sufficient context.
Finally, reliance on a template based or programmatic mapping layer as in Milan limits expressiveness and cannot support multiple joins, unions, filtering conditions, or normalization rules required for complex mapping scenarios. Unlike prior LLM-driven approaches we exploit the complementary benefits of symbolic search space reduction and LLM-guided (beyond syntactic or semantic) candidate matching and SQL view generation to achieve substantial improvements.  

% Second, without stage-specific supervision (high-quality few-shot exemplars), the model lacks guidance on the expected behavior for different mapping sub-types. Third, element-wise matching without appropriate contextualization (e.g., class context before property decisions) leads to locally plausible but globally inconsistent outputs. Fourth, the final mapping layer is template-based and cannot express joins, unions, filtering, and normalization rules required for complex mapping cases.

\textbf{RQ2: How does our symbolic decomposition-based LLM approach compare with traditional approaches?}
Our neuro-symbolic pipeline consistently outperforms traditional systems across all scenarios. In Table~\ref{tab:result}, our method achieves the best results on Conference-no-fk (0.79), Mondial (0.79), and NPD-Atomic (0.51), and substantially exceeds Milan, A4MO in average performance. The key reason is the complementary design of symbolic structure and neural reasoning. The symbolic layer decomposes the task into two stages: (i) matching (class, subclass, data property, object property) and (ii) mapping (SQL view generation with subject/object selection). This decomposition reduces task complexity and focuses each LLM call on a narrower, better-contextualized decision. 
% The neural layer (LLM) complements this structure by providing semantic or domain specific disambiguation for cryptic schema elements and generating expressive SQL for non-trivial mappings.
The neural layer (LLM) complements this symbolic structure by going beyond name similarity to provide semantic and domain-specific disambiguation for cryptic schema elements using ontology context, schema structure, and representative column values, while also generating expressive SQL for non-trivial mappings.
% The neural layer complements this symbolic structure by going beyond name similarity to provide semantic and domain-specific disambiguation for cryptic schema elements using ontology context, schema structure, and representative column values, while also generating expressive SQL for non-trivial mappings.

Ablation results in Table~\ref{tab:result} highlight the complementary roles of the matching and mapping stages. Dropping the matching stage substantially reduces performance, while replacing LLM-based mapping stage with programmatic mapping also results in a drop. These degradation indicate that both stages contribute essential, distinct capabilities to the pipeline, and their integration is critical for achieving high mapping quality.

\textbf{RQ3: How do prior few-shot-based LLM approaches perform compared with prior approaches?}
Few-shot supervision is beneficial, but its effectiveness depends on the quality and breadth of the demonstrations. As shown in Table~\ref{tab:result}, manually well-curated few-shots substantially improve performance over traditional and LLM-driven baselines, underscoring the importance of in-context learning for this task.
However, manually curating high-quality demonstrations is challenging and time-consuming at enterprise scale, requiring substantial domain expertise. 
As the number of ontology concepts and schema elements grows, identifying representative, diverse, and non-redundant few-shots becomes harder, making manual curation difficult to scale. 
% and a deep understanding of large ontologies and complex database schemas. 

We evaluate a prior automatic few-shot selection strategy proposed in Matchmaker, originally developed for schema-to-schema matching. When adapted to our pipeline, Matchmaker-style automatic few-shot selection yields limited, inconsistent improvements and continues to perform below the strongest traditional baseline, Milan.
A key limitation of automatically selected few-shots is their lack of coverage across the full range of mapping challenges. The selected examples tend to focus on straightforward semantic similarity cases, while failing to capture harder yet more consequential mapping patterns. In addition, Matchmaker’s selection strategy does not incorporate dependency-aware context across stages, further degrading the coherence and correctness of the demonstrations. As a result, the selected few-shots fail to convey the behaviors needed for difficult cases, leading to reduced robustness and lower overall mapping quality.

\textbf{RQ4: How does our dependency-aware, pattern-based self-demonstration approach compare with prior approaches?}
Our self-demonstration strategy addresses the above limitations by synthesizing stage-specific exemplars from reusable, domain-agnostic pattern families while respecting inter-stage dependencies. This produces demonstrations that are both diverse and structurally consistent with the pipeline's execution order.
As shown in Table~\ref{tab:result}, removing self-demonstrations lowers average performance to 0.38, underscoring the importance of high-quality supervision for the final outcome. 
An important and somewhat surprising observation is that, while the model performs poorly in zero-shot settings, enabling it to condition on self-generated few-shots guided by patterns and prior stage outputs leads to substantial performance gains.
With self-demonstrations enabled, our full approach achieves an average score of 0.56, establishing state-of-the-art performance on Conference-no-fk, Mondial, and NPD-atomic. 

Notably, our automatic self-demonstration strategy matches or slightly exceeds the performance of manually curated few-shots while fully eliminating manual curation overhead. The improvement is not due to simply increasing the number of examples, but due to generating the right demonstrations that are diverse, grounded in defined pattern families, and aligned with inter-stage dependencies in the pipeline.

\begin{wraptable}{r}{0.48\columnwidth}
  \vspace{-0.8\baselineskip}
  \centering
  \small
  \setlength{\tabcolsep}{4pt}
  \renewcommand{\arraystretch}{1.1}

  \begin{tabular}{@{}lccc@{}}
    \toprule
    \textbf{Scenario} & \textbf{Conf} & \textbf{Mondial} & \textbf{Avg} \\
    \midrule
    Table-to-KG~\cite{Vandemoortele2024ScalableTG}\# & 0.21 & 0.13 & 0.17 \\
    Magneto~\cite{Liu2024MagnetoCS}\#               & 0.49 & 0.34 & 0.40 \\
    \midrule
    \textbf{Our approach} & \textbf{0.56} & \textbf{0.65} & \textbf{0.61} \\
    \bottomrule
  \end{tabular}

  \caption{Results on RODI Benchmark with GPT4o. Conf: Conference-no-FK; \#: With programmatic mapping.}
  \label{tab:resultGPTo}
  \vspace{-0.8\baselineskip}
\end{wraptable}

\textbf{RQ5: Is our approach generalizable across LLMs?}
We perform a comparative analysis using GPT-4o
% \footnote{https://platform.openai.com/docs/models/gpt-4o} 
to assess the influence of LLM choice. The LLM-driven baselines are compared with our approach on the Conference-no-FK and Mondial scenarios. The NPD scenario, being significantly larger, is omitted from this specific evaluation due to budget limitations. Table \ref{tab:resultGPTo} presents these results, indicating that performance with GPT-4o is lower than Gemini-2.0-Flash (Table \ref{tab:result}). %, a difference that may stem from the prompts being extensively optimized for Gemini-2.0-Flash.
 With GPT-4o, our approach consistently surpasses Table-to-KG and Magneto across scenarios, demonstrating its generalizability across LLMs.

Overall, the answers to RQ1--RQ5 converge on a clear conclusion. Strong performance on DB-to-ontology mapping does not emerge from LLM capability alone, it depends on how the task is organized and supervised. In particular, symbolic decomposition improves tractability and dependency-aware pattern-guided self-demonstrations improve decision quality at each stage, and expressive mapping generation is essential for resolving structurally complex correspondences.\\
\textbf{Error Analysis}:
We analyze errors for the most complex NPD-Atomic scenario. Its 439 test query pairs evaluate mapping outcomes for distinct ontology concepts: 134 classes, 213 data properties, and 92 object properties. Our approach yields incorrect mappings for 46 classes, 157 data properties, and 63 object properties. We randomly sample and analyze 20 test queries for each ontology concept.

For classes, 12 errors stem from incorrect matches, 6 from incorrect SQL views, and 2 from incorrect subject/object column selection. Incorrect matches mainly arise when a class corresponds to multiple tables or when predicted and actual tables are ambiguous. For example, the \texttt{Jacket4LegsFacility} class matches both \texttt{facility\_fixed} and \texttt{facility\_moveable} tables (where \texttt{fclkind} = `\texttt{JACKET 4 LEGS}'), but our approach identifies only \texttt{facility\_fixed}, relying on the definition of `Jacket' as a fixed offshore structure anchored to the seabed. Similarly, \texttt{ParcellBAA} class is incorrectly mapped to \texttt{baaarea} table instead of \texttt{bsns\_arr\_area} table, because both tables share similar column names and contain \texttt{baakind} column with value \texttt{`PARCELL'}, making them hard to distinguish. SQL-view errors involve (i) incorrect or missing categorical filter values and (ii) incorrect column selection. For example, although \texttt{Onshorefacility} class is correctly matched to column \texttt{fclkind}, the generated SQL filters by \texttt{`LANDFALL'} instead of \texttt{`ONSHORE FACILITY'}. Likewise, for \texttt{LithostratigraphicUnit} class, correctly matched to \texttt{wellbore\_formation\_top} table, the SQL view selects \texttt{lsuname} and \texttt{lsulevel} columns instead of all primary keys of the table.

For data properties, 10 errors are due to incorrect matches and 10 to incorrect SQL view generation. 
% Most incorrect-match errors stem from wrongly inferred domain classes during data enrichment. 
Most incorrect‑match errors stem from incorrect domain class inferences during the prior metadata inference step.
For example, the data property \texttt{sourcePressure} is inferred to belong to \texttt{SeismicSurvey} class. This inference is semantically plausible, since source pressure relates to seismic acquisition, but the ground-truth mapping assigns it to \texttt{SurveyArea} and \texttt{SurveyMultilineArea}. These more specific domain classes are difficult to infer from the property name alone. SQL-generation errors involve incorrect column selection or missing information. For example, although \texttt{dateStatusTo} data property is correctly matched to the \texttt{field\_activity\_status\_hst} column, the generated SQL view does not filter out the placeholder date \texttt{`9999-12-31T00:00:00'}, leading to errors. A particularly challenging case is \texttt{coreIntervalTop} property of \texttt{WellboreCore} class, correctly matched to \texttt{wlbCoreIntervalTop} column in \texttt{wellbore\_core} table. The ground-truth SQL view is as follows:
\begin{lstlisting} [
          language=SQL,
          showspaces=false,
          basicstyle=\ttfamily,
        ]
(SELECT wellbore_core_id, wlbnpdidwellbore, 
wlbcorenumber, wlbcoreintervaltop FROM 
wellbore_core  WHERE wlbcoreintervaluom ='[m]') 
UNION ALL (SELECT wellbore_core_id,
wlbnpdidwellbore, wlbcorenumber, 
wlbcoreintervaltop * 0.3048  FROM wellbore_core 
WHERE wlbcoreintervaluom ='[ft]')
\end{lstlisting}
The ontology does not specify that all values must be normalized to meters using \texttt{wlbCoreIntervalUom} column, making the correct SQL view impossible for the LLM to generate.

For object properties, 16 errors are due to incorrect matches and 4 to incorrect SQL views. Among incorrect-match errors, 6 stem from wrongly inferred domain/range classes and 3 from incorrect domain/range matches. SQL-view errors usually involve incorrect column selection or missing information. For example, the LLM fails to correctly map the object property \texttt{productionForField} because the reason for excluding value `44576' is not specified. The ground-truth SQL view is:
 \begin{lstlisting} [
          language=SQL,
          showspaces=false,
          basicstyle=\ttfamily,
        ]
SELECT prfnpdidinformationcarrier, prfyear
FROM field_production_yearly 
WHERE prfnpdidinformationcarrier <> `44576'.
\end{lstlisting}

\section{Conclusion}

Enterprise-scale schema-to-ontology mapping remains a challenging problem due to severe semantic heterogeneity, noisy and incomplete metadata, cryptically named schema elements, and the fundamental mismatch in abstraction between relational data and ontological models. This work shows that while LLMs offer strong semantic reasoning capabilities, they are ineffective when applied in a naïve or weakly structured manner. Our study establishes that a neuro-symbolic approach is essential for achieving reliable and accurate mappings. By decomposing the task into structured stages and using symbolic constraints to systematically narrow the search space, the approach improves tractability and enables LLMs to operate within focused contexts leading to improved accuracy.

At the same time, decomposition alone is insufficient to fully address the challenges of schema-to-ontology mapping. An important finding of this work is the surprising effectiveness of self-demonstrations in supervising each stage of the pipeline. Self-demonstrations refer to automatically generated in-context examples produced by the language model itself, showing how specific matching or mapping decisions should be made for a given sub-task. These examples are guided by reusable, domain-agnostic mapping patterns, ensuring coverage of common and complex correspondence behaviors observed in real integration settings. By tailoring self-demonstrations to individual sub-tasks, the model receives explicit and diverse guidance that clarifies expected reasoning and outputs. Furthermore, generating these examples in a dependency-aware manner, where downstream stages are conditioned on the outputs of earlier ones, preserves consistency across stages and limits error propagation. Together, these properties enable more reliable local decisions, which accumulate to produce more accurate mappings.
Leveraging the neuro-symbolic decomposition together with pattern-guided context-aware self-demonstrations, our method establishes state-of-the-art performance on the three most challenging scenarios from the RODI benchmark, demonstrating $\sim$25 percentage points  improvements in the F1 score over  prior automated baselines.

% \textcolor{blue}{
As future work, we plan to reduce cost by partially replacing LLM-based reasoning with programmatic and template-driven mapping for well-structured and repetitive correspondence patterns. In addition, instead of producing a single mapping decision, we will output ranked candidate matches accompanied by confidence scores, enabling expert‑in‑the‑loop workflows where domain experts can efficiently validate, adjust, or correct mappings by inspecting a small set of plausible alternatives rather than performing exhaustive manual verification.
% }

\section*{Declaration on Generative AI}
During the preparation of this work, the author(s) used Microsoft Copilot in order to paraphrase and reword portions of human-authored text. After using this tool/service, the author(s) reviewed and edited the content as needed and take(s) full responsibility for the publication's content.

\noindent\textbf{Tools and services:} Microsoft Copilot.

\noindent\textbf{Tools' contributions:} Paraphrasing and rewording (linguistic refinement) of human-authored text.

\bibliographystyle{vancouver}
\bibliography{sample-base}

\end{document}